\documentclass[letterpaper]{article}
\usepackage{uai2020}
\usepackage[margin=1in]{geometry}

\usepackage{times}

\usepackage[style=authoryear,firstinits=true,maxnames=2,backend=bibtex,natbib,doi=false,isbn=false,url=false,date=year]{biblatex}
\defbibheading{uai}{\subsubsection*{References}}

\usepackage{algorithm, algorithmic} 
\usepackage{booktabs}
\usepackage{amsmath, amssymb}

\usepackage{causality}
\usepackage{siunitx}

\usepackage{tikz}
\usetikzlibrary{arrows, shapes.misc}

\title{MASSIVE: Tractable and Robust Bayesian Learning of Many-Dimensional Instrumental Variable Models}

\author{ Ioan Gabriel Bucur \And 
	Tom Claassen \\
	Department of Data Science \\
	Institute for Computing and Information Sciences \\
	Radboud University \\
	Nijmegen, The Netherlands \\
	\And Tom Heskes 
}

\begin{document}
	
	\maketitle
	
	\begin{abstract}
		The recent availability of huge, many-dimensional data sets, like those arising from genome-wide association studies (GWAS), provides many opportunities for strengthening causal inference. One popular approach is to utilize these many-dimensional measurements as instrumental variables (instruments) for improving the causal effect estimate between other pairs of variables. Unfortunately, searching for proper instruments in a many-dimensional set of candidates is a daunting task due to the intractable model space and the fact that we cannot directly test which of these candidates are valid, so most existing search methods either rely on overly stringent modeling assumptions or fail to capture the inherent model uncertainty in the selection process. We show that, as long as at least some of the candidates are (close to) valid, without knowing a priori which ones, they collectively still pose enough restrictions on the target interaction to obtain a reliable causal effect estimate. We propose a general and efficient causal inference algorithm that accounts for model uncertainty by performing Bayesian model averaging over the most promising many-dimensional instrumental variable models, while at the same time employing weaker assumptions regarding the data generating process. We showcase the efficiency, robustness and predictive performance of our algorithm through experimental results on both simulated and real-world data.
	\end{abstract}

	\section{INTRODUCTION}
	
	Causal inference is a fundamental topic of research in the biomedical sciences, where the relationship between an exposure to a putative risk factor and a disease outcome or marker is often studied. The gold standard for answering causal questions -- e.g., does an intake of vitamin D supplements reduce the risk of developing schizophrenia? -- is to perform a \emph{randomized controlled trial} (RCT), in which the exposure (treatment) is assigned randomly to the participants. The purpose of randomization is to eliminate potential confounding due to variables influencing both the exposure and the outcome. Unfortunately, performing an RCT is often unfeasible due to monetary, ethical, or practical constraints~\citep{benson_comparison_2000}. On the other side of the fence, there are vast amounts of medical data available from observational studies, but estimating a causal effect from such data is prone to confounding, reverse causation, and other biases~\citep{sheehan_mendelian_2008}. 
	
	With the advent of high-throughput genomics, an enormous amount of observational genetic data has been collected in large-scale \emph{genome-wide association studies} (GWAS). There is great potential in using this genetic information for strengthening causal inference in observational designs, where the causal effect is obfuscated by potentially unmeasured confounding~\citep{visscher_10_2017}. One popular and powerful systematic approach that can be exploited is to make use of so-called \emph{instrumental variables} or \emph{instruments} ~\citep{angrist_identification_1996}. In recent years, instrumental variable analysis has become prevalent in the field of genetic epidemiology under the moniker \textit{Mendelian randomization}. Mendelian randomization (MR) refers to the random segregation and assortment of genes from parent to offspring, as stated by Mendel's laws, which can be seen as analogous to the randomization induced in an RCT~\citep{hingorani_natures_2005}. In MR studies, \emph{genetic variants}, such as the allele at a particular location in the genome, fulfill the role of instruments~\citep{lawlor_mendelian_2008}. For example, a gene encoding a major enzyme for alcohol metabolism (ALDH2) has been used as a proxy measure for alcohol consumption with the goal of investigating the latter's effect on the risk of coronary heart disease~\citep{davey_smith_mendelian_2014}. 
	
	\begin{figure}[!htb]
		\centering
		
		\begin{tikzpicture}[shorten >=1pt, auto, node distance = 2.5cm,
			semithick]
			
			\tikzset{vertex/.style = {shape = circle, draw, minimum size=0.5cm}}
			\tikzset{edge/.style = {->,> = latex'}}
			
			\node[vertex] (1) at (0, 0) {$\iv$};
			\node[vertex] (2) at (3, 0) {$\exs$};
			\node[vertex] (3) at (6, 0) {$\out$};
			\node[vertex, dotted] (4) at (4.5, 2) {$\conf$};
			
			\draw[edge] (1) to (2);
			\draw[edge, ultra thick] (2) to (3);
			\draw[edge] (4) to (2);
			\draw[edge] (4) to (3);
			\node (23) at (4.5, 0.25) {$\ce$};
			\node (42) at (3.5, 1.25) {$\cc_\exs$};
			\node (43) at (5.5, 1.25) {$\cc_\out$};
			
		\end{tikzpicture}
		
		\caption{Graphical description of the causal model assumed in instrumental variable analyses. In the figure above, $\exs$ is the exposure, $\out$ is the outcome variable, $\iv$ is the instrument, and $\conf$ represents potentially unmeasured confounding. Note that the association between $\iv$ and $\exs$ need not be causal, but we can assume it here  for simplicity without losing any generality.}  \label{fig:iv_model}
	\end{figure}
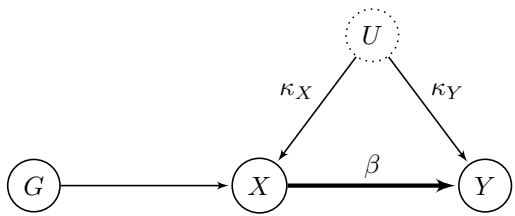
	
	Formally, an instrumental variable (IV) is a third variable in regression analysis that is correlated with both exposure and outcome, but affects the outcome only through its association with the exposure. A \emph{valid} instrument follows the causal model depicted in Figure~\ref{fig:iv_model}. An IV thus acts as a proxy for the exposure that is not susceptible to the same degree of confounding. A key challenge in instrumental variable methods is finding the right instrument(s) for performing the analysis~\citep{john_assessing_2019}. Due to the unmeasured confounding of the $\exs-\out$ association, this model cannot be elucidated from observed data unless we are willing to make strong assumptions about the generating process~\citep{cornia_type-ii_2014, silva_learning_2017}.
	
	Genetic variants are particularly suitable as candidate instrumental variables, since they are fixed at conception and more robust against confounding due to environmental factors~\citep{davey_smith_clustered_2007}. Nevertheless, the validity of genetic instruments is also not easily testable from data. To make matters worse, many genes affect multiple traits, meaning that the outcome variable $\out$ could be influenced by $\iv$ via different causal pathways. This violation of the instrumental variable assumptions is known in the Mendelian randomization literature as \emph{horizontal pleiotropy}~\citep{chesmore_ubiquity_2018}. Horizontal pleiotropy, to which we will refer from now simply as \emph{pleiotropy}, is usually shown as a directed arrow from genetic variant ($\iv$) to outcome ($\out$) and the implied (direct) causal effect from $\iv$ to $\out$ is called a \emph{pleiotropic effect}.
	
	Searching for instruments in a haystack of potentially relevant genetic variants with unknown biological function is akin to a many-dimensional variable selection problem. To solve this problem, we adopt a \emph{spike-and-slab} prior on the pleiotropic effects ($\iv \to \out$) to encourage sparse solutions through \emph{selective shrinkage}~\citep{ishwaran_spike_2005}. The `spike' captures the prior distribution of coefficients that are close to zero, corresponding to valid instruments, while the `slab' models the prior distribution of coefficients that are significantly different from zero. Even though we do not know \emph{a priori} which of the genetic variants are (close to being) valid instruments, by using the \emph{wisdom of the crowd}~\citep{surowiecki_wisdom_2005}, where the crowd is the many-dimensional set of potential candidates, we are able to separate the wheat from the chaff, as we will later see in Section~\ref{sec:empirical}. We show that, as long as there are at least some valid instruments to be found in the haystack, the causal effect of interest can be reasonably estimated by using the proposed prior.
	
	In this work, we consider a general Bayesian causal model subsuming the IV model in which a large number of (genetic) covariates have the potential to act as instrumental variables. We assume a hierarchical discrete scale mixture (spike-and-slab) prior on the pleiotropic effects to consider every possible combination of valid and invalid instruments. We then introduce an algorithm (\alg{MASSIVE}) which we use to perform Bayesian model averaging (BMA) over this mixture space so as to properly handle the uncertainty in choosing the covariates to be used as instruments. The algorithm features two components: (1) a Markov Chain Monte Carlo Model Composition (\alg{MC3}) stochastic search procedure~\citep{madigan_bayesian_1995} and (2) an approximation procedure based on \emph{Laplace's method}~\citep{bishop_pattern_2006} for determining the model evidence (marginal likelihood). We show the robustness and tractability of our approach in both simulated studies and real-world examples.

	\section{RELATED WORK}
	
	A number of methods have been suggested for selecting instrumental variables out of a rich set of candidates. \citet{swerdlow_selecting_2016} have outlined a set of principles for selecting instruments in MR analyses using a combination of statistical criteria and relevant biological  knowledge. \citet{belloni_sparse_2012}, on the other hand, have proposed a data-driven approach for model selection based on Lasso methods. \citet{agakov_sparse_2010} have built an approach for extracting the most reliable instruments by using approximate Bayesian inference with sparseness-inducing priors on linear latent variable models. Finally,~\citet{berzuini_bayesian_2020} have developed a Bayesian solution in which the horseshoe shrinkage prior is imposed on potential pleiotropic effects. These methods, however, are designed to select the most likely IV model and do not account for potential model uncertainty. Moreover, some of these methods require individual patient data, which is often unavailable, as input.
	
	A number of model averaging solutions have also been proposed.~\citet{eicher_bayesian_2009} have used BMA to average over the set of potential models in the first stage of two-stage least squares (2SLS), which means that the selection of instruments is based on the strength of their association with the exposure. The model evidences are approximated using the \textit{Bayesian information criterion}~\citep{schwarz_estimating_1978}. \citet{eicher_bayesian_2009} later extended their approach in \citep{lenkoski_two-stage_2014} by also accounting for model uncertainty in the second stage of 2SLS. In a similar vein,~\citet{karl_instrumental_2012} developed the \alg{IVBMA} algorithm to incorporate model uncertainty into IV estimation by exploring the model space using stochastic search guided by analytically derived conditional Bayes factors. More recently, \citet{shapland_bayesian_2019} have proposed using the  \alg{IVBMA} approach for Mendelian randomization with dependent instruments. The above-mentioned methods, however, work under the assumption that the chosen candidates are all valid instruments. This means that the algorithms are no longer consistent if any of the IV assumptions are violated.
	
	\citet{gkatzionis_bayesian_2019} have introduced a comparable Bayesian model averaging method (\alg{JAM-MR}) in which genetic variants likely to exhibit horizontal pleiotropy, thereby violating the IV assumptions, are penalized via a pleiotropic-loss function. \alg{JAM-MR} implements a standard reversible-jump MCMC stochastic search scheme for exploring the model space.  However, the estimated causal effect for each model is obtained using the classical \emph{inverse-variance weighted} (IVW) estimator~\citep{burgess_mendelian_2015}, meaning that there is no complete description of the parameter uncertainty.

	\section{MODEL}
	
	Currently no published method offers a complete Bayesian solution for handling both the uncertainty in selecting the most promising candidates out of a many-dimensional set of potential instruments and the uncertainty in estimating the causal effect using those instruments. We propose to address this shortcoming with our \alg{MASSIVE} (Model Assessment and Stochastic Search for Instrumental Variable Estimation) Bayesian approach, which is designed to reliably estimate the studied causal effect as long as at least one of the candidate instruments is close to valid. This condition is weaker than causal assumptions typically made in related work, e.g., a plurality of the candidate instruments are valid (the most common pleiotropic effect is zero) or the pleiotropic effects are balanced (on average they cancel each other out).
	
	Our method incorporates Bayesian model averaging to further relax the IV causal assumptions by searching for the most plausible many-dimensional IV models, thereby properly accounting for uncertainty in the model selection. Our algorithm provides as output a posterior distribution over the causal effect that appropriately reflects the uncertainty in the estimate, as well as posterior inclusion probabilities indicating which candidates are likely to be valid instruments. Finally, our approach does not rely on having access to individual-level data, and instead can use publicly available summary data from large-scale GWAS as input. This constitutes a significant practical advantage, as access to information about individuals is often restricted, for instance due to privacy concerns~\citep{pasaniuc_dissecting_2017}. 
	
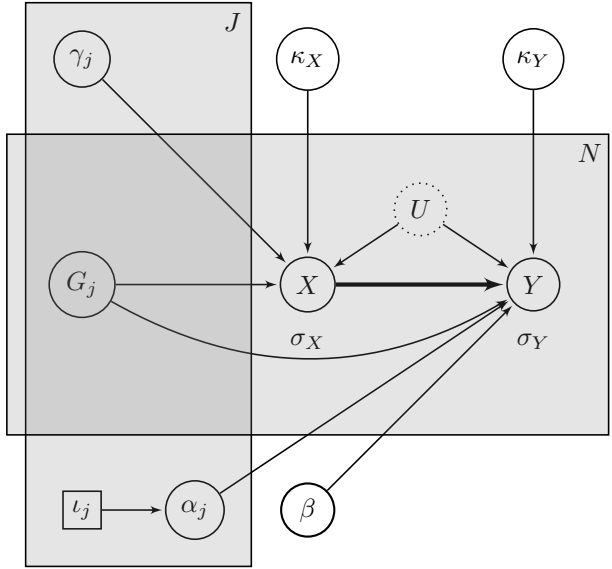
\begin{figure}[!htb]
		\centering
		
		\begin{tikzpicture}[shorten >=1pt, auto, node distance = 2.5cm,
			semithick]

			\tikzset{vertex/.style = {shape = circle, draw, minimum size=0.5cm}}
			\tikzset{param/.style = {shape = circle, draw, minimum size=0.5cm}}
			\tikzset{hyper/.style = {shape = rectangle, draw, minimum size=0.5cm}}
			\tikzset{edge/.style = {->,> = latex'}}
			
			\node[vertex] (1) at (0, 0) {$\iv_j$};
			\node[vertex] (2) at (3, 0) {$\exs$};
			\node[vertex] (3) at (6, 0) {$\out$};
			\node[vertex, dotted] (4) at (4.5, 1) {$\conf$};
			
			\node[param] (12) at (0, 3) {$\is_j$};
			\node[param] (13) at (1.5, -3) {$\ply_j$};
			\node[hyper] (130) at (0, -3) {$\li_j$};
			\node[param] (42) at (3, 3) {$\cc_\exs$};
			\node[param] (43) at (6, 3) {$\cc_\out$};
			\node[param, thick] (23) at (3, -3) {$\mathbf{\ce}$};
			\node (20) at (3, -0.75) {$\sd_\exs$};
			\node (30) at (6, -0.75) {$\sd_\out$};

			\draw[edge] (1) to (2);
			\draw[edge, ultra thick] (2) to (3);
			\draw[edge] (4) to (2);
			\draw[edge] (4) to (3);
			\draw[edge] (1) to[bend right] (3);
			\draw[edge] (12) to (2);
			\draw[edge] (13) to (3);
			\draw[edge] (42) to (2);
			\draw[edge] (23) to (3);
			\draw[edge] (43) to (3);
			\draw[edge] (130) to (13);
			
			\node at (2, 3.5) {$J$};
			\node at (6.75, 1.75) {$N$};
			
			\draw [fill=gray, fill opacity=0.2] (-0.75, -3.75) rectangle (2.25, 3.75);
			\draw [fill=gray, fill opacity=0.2] (-1, -2) rectangle (7, 2);
			
		\end{tikzpicture}
		
		\caption{Graphical description of our assumed generative model. We denote the exposure variable by $\exs$ and the outcome variable by $\out$. We are interested in the causal effect from $\exs$ to $\out$, which is denoted by $\ce$. The association between $\exs$ and $\out$ is obfuscated by the unobserved variable $\conf$, which we use to model unmeasured confounding explicitly. The shaded plate indicates replication across the $J$ independent genetic variants $\iv_j, j \in \{1, 2, ..., J\}$. Note that the replication also applies to the parameters $\is_j$ and $\ply_j$.}  \label{fig:model}
	\end{figure}

	In our model, we assume that the data is generated from the following \emph{structural equation model}~\citep{bollen_structural_1989}:
	\begin{equation}
		\label{eqn:sem}
		\begin{aligned}
			\conf &:= \err_\conf \\
			\iv_j &:= \err_{\iv_j} \\
			\exs &:= \sum_j \is_j \iv_j + \cc_\exs \conf + \err_\exs \\ 
			\out &:= \sum_j \ply_j \iv_j + \cc_\out \conf + \ce \exs + \err_\out
		\end{aligned}.
	\end{equation}
	The associated generating model is depicted graphically in Figure~\ref{fig:model}. We are interested in estimating the (linear) causal effect from exposure ($\exs$) to outcome ($\out$), denoted by $\ce$. To aid estimation, we have measurements from $\niv$ covariates, denoted by $\iv_j$, at our disposal. Each covariate is associated in the model with both the exposure $\exs$, via the $\is_j$ parameters, and the outcome $\out$, via the $\ply_j$ parameters. Finally, the unmeasured confounding is characterized by the coefficients $\cc_\exs$ and $\cc_\out$. 
	
	We assume that the noise terms of $\exs$, $\out$, and the unmeasured confounder $\conf$ are normally distributed. We can assume without loss of generality that $\err_\conf \sim \normd(0, 1)$ by appropriately rescaling the confounding coefficients. The exposure and outcome terms are normally distributed with unknown scale parameters, i.e., $\err_\exs \sim \normd(0, \sd_\exs^2)$ and $\err_\out \sim \normd(0, \sd_\out^2)$. The random vector $(\exs, \out) \given \ivv$ then follows the Conditional Gaussian distribution (\emph{CG-distribution} in~\citep{lauritzen_graphical_1989}):
	\begin{equation*} \label{eqn:cond_gauss}
		\left. \begin{bmatrix} \exs \\ \out \end{bmatrix} \right| \ivv \sim \normd(\bmu(\ivv), \bSigma),
	\end{equation*}
	where $\bmu(\ivv) = \begin{bmatrix} \isv & \ce \isv + \plyv \end{bmatrix}^\trp \ivv$ and $\bSigma =$
	{\small
		$$
		=\begin{bmatrix} \sd_\exs^2 + \cc_\exs^2 & \ce (\sd_\exs^2 + \cc_\exs^2) + \cc_\exs \cc_\out \\ \ce (\sd_\exs^2 + \cc_\exs^2) + \cc_\exs \cc_\out & \sd_\out^2 + \ce^2 \sd_\exs^2 + (\cc_\out + \ce \cc_\exs)^2 \end{bmatrix}.
		$$
	}%
	
	We now assume that $\nobs$ independent and identically distributed observations $\data = (\ivv_i, \exs_i, \out_i)_{1 \le i \le \nobs}$ are drawn from the structural equation model described in~\eqref{eqn:sem}. The conditional Gaussian observed data likelihood reads
	{\small
		\begin{equation}
			\label{eqn:likelihood}
			\lik \left( \left. \begin{bmatrix} \exs \\ \out \end{bmatrix} \right| \ivv \right) = (4\pi^2|\bSigma|)^{-\frac{\nobs}{2}} \exp\left\{-  \frac{\nobs}{2} \textrm{tr}(\bSigma^{-1} \bfS) \right\}, 
		\end{equation}
		with $\bfS = \by{\nobs} \sum\limits_{i = 1}^\nobs \left\{\begin{bmatrix} \exs_i \\ \out_i \end{bmatrix} - \bmu(\ivv_i) \right\} \left\{\begin{bmatrix} \exs_i \\ \out_i \end{bmatrix} - \bmu(\ivv_i) \right\}^\trp$.
	}

	\subsection{PRIORS}
	
	In order to avoid any scaling issues, we first divide each structural equation in~\eqref{eqn:sem} by the scale of the noise term. We then define priors on the scale-free interactions. The scaled structural parameters are
	\begin{align*}
		\sis_j &= \sd_{\iv_j} \sd_\exs^{-1} \is_j; \quad \sply_j = \sd_{\iv_j} \sd_\out^{-1} \ply_j; \\
		\sce &= \sd_\exs \sd_\out^{-1} \ce; \quad \scc_\exs = \sd_\exs^{-1} \cc_\exs; \quad \scc_\out = \sd_\out^{-1} \cc_\out.
	\end{align*}
	For each scaled pleiotropic effect ($\sply_j$), we propose a scale mixture of two normal distributions~\citep{ishwaran_spike_2005}, where the scale is determined by the value of a latent indicator variable $\li_j$. The component with lower (higher) variance encompasses our prior belief that the pleiotropic effect is a priori `weak' / irrelevant (`strong' / relevant). This hierarchical prior is identical to the one proposed by~\citet{george_variable_1993} for their Stochastic Search Variable Selection (\alg{SSVS}) algorithm. 
	
	The standard deviation of the `spike' (lower variance) component and of the `slab' (higher variance) component can be set based on our prior knowledge or assumptions regarding the size of relevant and irrelevant parameters. For example,~\citet{george_variable_1993} have proposed a semiautomatic approach for selecting the spike-and-slab hyperparameters based on the intersection point of the two mixture components and the relative heights of the component densities at zero. For the more general situation when prior knowledge is not available, we propose a simple empirical approach for choosing these hyperparameters starting from the belief (assumption) that the measured interactions between $\ivv$ and $\exs$ are all relevant, which we can expect in most analyses since the first criterion by which potential instruments are chosen is the relevance of their association with the exposure. We describe the procedure for empirically determining prior hyperparameters in the supplement.
	
	For the scaled instrument strengths $\sis_j$, we propose a normal prior with the same variance as the slab component, under the mild assumptions that genetic interactions with different traits are of the same size and that the instrument strengths correspond are strong (relevant) interactions. For the causal effect ($\sce$) and the confounding coefficients ($\scc_\exs$ and $\scc_\out$), we choose a very weakly informative normal prior proposed by~\citet{gelman_prior_2020}. For the scale parameters ($\sd_\exs$ and $\sd_\out$), we propose an improper uniform prior on the log-scale, corresponding to Jeffreys's scale-invariant prior~\citep{gelman_bayesian_2013}. The final Bayesian generating model is
	\begin{equation} \label{eqn:bayesian_gen}
		\begin{aligned}
			\li_j &\sim \bernd(0.5); \\
			\sply_j &\sim \li_j \cdot \normd(0, \sd_\slab^2) + (1 - \li_j) \cdot \normd(0, \sd_\spike^2); \\
			\sis_j &\sim \normd(0, \sd_\slab^2); \quad \sce \sim \normd(0, 10); \\
			\scc_\exs &\sim \normd(0, 10); \quad \scc_\out \sim \normd(0, 10);  \\
			p&(\log \sd_\exs) \propto 1; p(\log \sd_\out) \propto 1; \\
			\left. \begin{bmatrix} X \\ Y \end{bmatrix} \right| \ivv &\sim \normd\left( \begin{bmatrix} \isv^\trp \ivv \\ (\beta \isv + \plyv)^\trp \ivv \end{bmatrix}, \bSigma \right).
		\end{aligned}
	\end{equation}
	
	\subsection{BAYESIAN MODEL AVERAGING}
	
	In our approach we use the general framework of Bayesian Model Averaging to incorporate the uncertainty in instrument candidate validity by combining the causal effect estimates from reasonable instrument combinations. Instead of relying on a single model for estimating our causal effect $\ce$, we average the estimates over a number ($K$) of promising models, weighing each result by the model posterior
	$$ p(\ce \given \data) = \sum_{k=1}^K p(\ce \given M_k, \data) p(M_k \given \data).$$
	Our assumed generating model in~\eqref{eqn:bayesian_gen} has $2 \niv + 5$ parameters, $\parm = (\sisv, \splyv, \sce, \scc_\exs, \scc_\out, \log\sd_\exs, \log\sd_\out)$, where $\niv$ is the number of candidates. There are $\niv$ latent indicator variables $\li_j$ corresponding to the parameters $\sply_j$ which indicate whether each parameter is `weak' (generated by the `spike' component) or `strong' (generated by the `slab' component). The full multivariate prior thus is a mixture of $K = 2^\niv$ multivariate Gaussian priors (the uniform prior on the log-scale parameters can be seen as a limiting case of a Gaussian prior). We refer to each mixture component as a different \emph{model}. The difference between these models lies solely in the prior beliefs we assume on the pleiotropic effect strengths.
	
	It is intractable to consider the entire space of $2^\niv$ models (multivariate indicator instances), so we instead search for a subset that best fits the data using MCMC Model Composition (\alg{MC3})~\citep{madigan_bayesian_1995}. If an unspecified subset of the $\niv$ candidates are close to being valid instruments, then only a small number of models will be a good fit to the data. We can thus obtain a good approximation of the model posterior probabilities without averaging over the entire model space. The idea of \alg{MC3} is to construct a Markov chain that moves through the class of models $\model$ = $\{0, 1\}^{\niv}$. For each model $M$ we define a neighborhood consisting of the $\niv$ models that have only one indicator variable different than $M$, and we allow transitions only into the set of neighbors, with equal probability. A new model $M'$ in the neighborhood is then accepted with probability
	\begin{equation*}
		\min \left\{ 1, \frac{p(M' \given \data)}{p(M \given \data)} \right\}, 
	\end{equation*}
	where $p(M \given \data)$ is the posterior probability of model $M$. The posterior probability is given by Bayes's theorem
	$$ p(M \given \data) = \frac{p(\data \given M) p(M)}{\sum p(\data \given M') p(M')},$$
	where
	$$ p(\data \given M) = \int_{\parm} p(\data \given \parm, M) p(\parm \given M) \diff\parm$$ is the model evidence. Here, the latent indicators $\li_j$ are part of the model definition and their choice determines the parameter prior given the model, i.e., $p(\parm \given M)$. As prior over the model space, we consider the simple uniform prior $p(M) = 2^{-\niv}$. This prior corresponds to the assumption that each parameter is as likely to be `relevant' as `irrelevant' a priori, i.e., $\li_j \sim \bernd(0.5)$ in~\eqref{eqn:bayesian_gen}. Other priors on the model space could be easily accommodated to indicate a prior belief in the presence or absence of pleiotropic effects. 
	
	A key challenge when considering a general approach such as the one proposed here is estimating the \emph{evidence} (\emph{marginal likelihood}) for each model. Since the integral is not analytically tractable for the proposed likelihood and priors, we have to resort to approximation methods. One idea would be to approximate the evidence with a \emph{nested sampling} algorithm~\citep{skilling_nested_2006}, but this procedure is relatively slow, so we instead propose to approximate the evidence more efficiently using Laplace's method, similar to~\citet{rue_approximate_2009}.

	\section{ALGORITHM}
	
	\subsection{FINDING THE POSTERIOR OPTIMA}
	
	When sampling a certain combination of indicator variables, we need to compute the corresponding approximate model evidence using Laplace's method. We need to find local posterior optima over the $2\niv+5$ parameters $\sparm = (\sisv, \splyv, \sce, \log\sd_\exs, \log\sd_\out, \scc_\exs, \scc_\out)$. Despite the simplicity of our chosen priors, we are dealing with a many-dimensional multimodal optimization problem. We tackle the issue by first separating our model parameters into those pertaining to observed variables, denoted by $\sobsm = (\sisv, \splyv, \sce, \log\sd_\exs, \log\sd_\out)$, and those pertaining to the unobserved variable, denoted by $\scnfm = (\scc_\exs, \scc_\out)$. 
	
	To guide the optimization, we use the fact that for each value of the confounding coefficients in $\scnfm$, we can analytically derive the maximum likelihood estimate for $\sobsm$. For the details of deriving the ML estimate, please see the supplement. Thus, if we attempt to perform inference via maximum likelihood estimation, we arrive at a two-dimensional manifold of equally good solutions for the equation system. We propose to start the posterior optimization procedure from the bivariate ML manifold, for each considered model. We develop a smart procedure for choosing starting points on the manifold, described in the supplement, in which we look for (sparse) parameter combinations where some of the parameters are close to zero. The optimization initialization list $\mathcal{L}$ is given as input to the posterior approximation in Algorithm~\ref{alg:la}.
	
	By analyzing the optimization results in the $\scnfm$ space, we have identified at most five local optima for each model. Note that these optima constitute pairs that are symmetric with respect to the origin. This is because the value of the posterior does not change if we replace $(\scc_\exs, \scc_\out)$ with $(-\scc_\exs, -\scc_\out)$. One possible optimum occurs at the critical point corresponding to the \emph{no confounding} scenario, when the confounding coefficients are close to zero. We can find this optimum efficiently, if it exists, by starting the posterior optimization from the maximum likelihood parameters obtained when setting $\scc_\exs = \scc_\out = 0$. 
	
	\subsection{COMPUTING THE APPROXIMATION}
	
	\begin{algorithm}[!htb]
		\caption{\alg{Approximate Posterior}}
		\label{alg:la}
		\begin{algorithmic}
			\STATE {\bfseries Input:} data $\bfZ = [\ivv_i, \exs_i, \out_i]_{1 \le i \le \nobs}$, model $M$, optimization initialization list $\mathcal{L}$ 
			\FOR{$(\scc_\exs, \scc_\out)$ in $\mathcal{L}$}
			\STATE $\sparm^\ML$ = \alg{get\_ML\_estimate}($\bfZ, \scc_\exs, \scc_\out)$	
			\STATE $\sparm^\MAP$ = \alg{optimize}($posterior(\bfZ, M), \sparm^\ML$)
			\STATE LA = \alg{Laplace\_approximation}($\sparm^\MAP$)
			\STATE \textbf{Save:} $\sparm^\MAP(\scc_\exs, \scc_\out)$, LA$(\sparm^\MAP)$
			\ENDFOR
			\STATE Eliminate potential duplicates from optima list;
			\STATE Compute total model evidence from LA list;
			\STATE {\bfseries Output:} Mixture of LA($\sparm^\MAP$), model evidence
		\end{algorithmic}
	\end{algorithm}
	
\begin{figure}[!htb]
		\includegraphics[width = \linewidth]{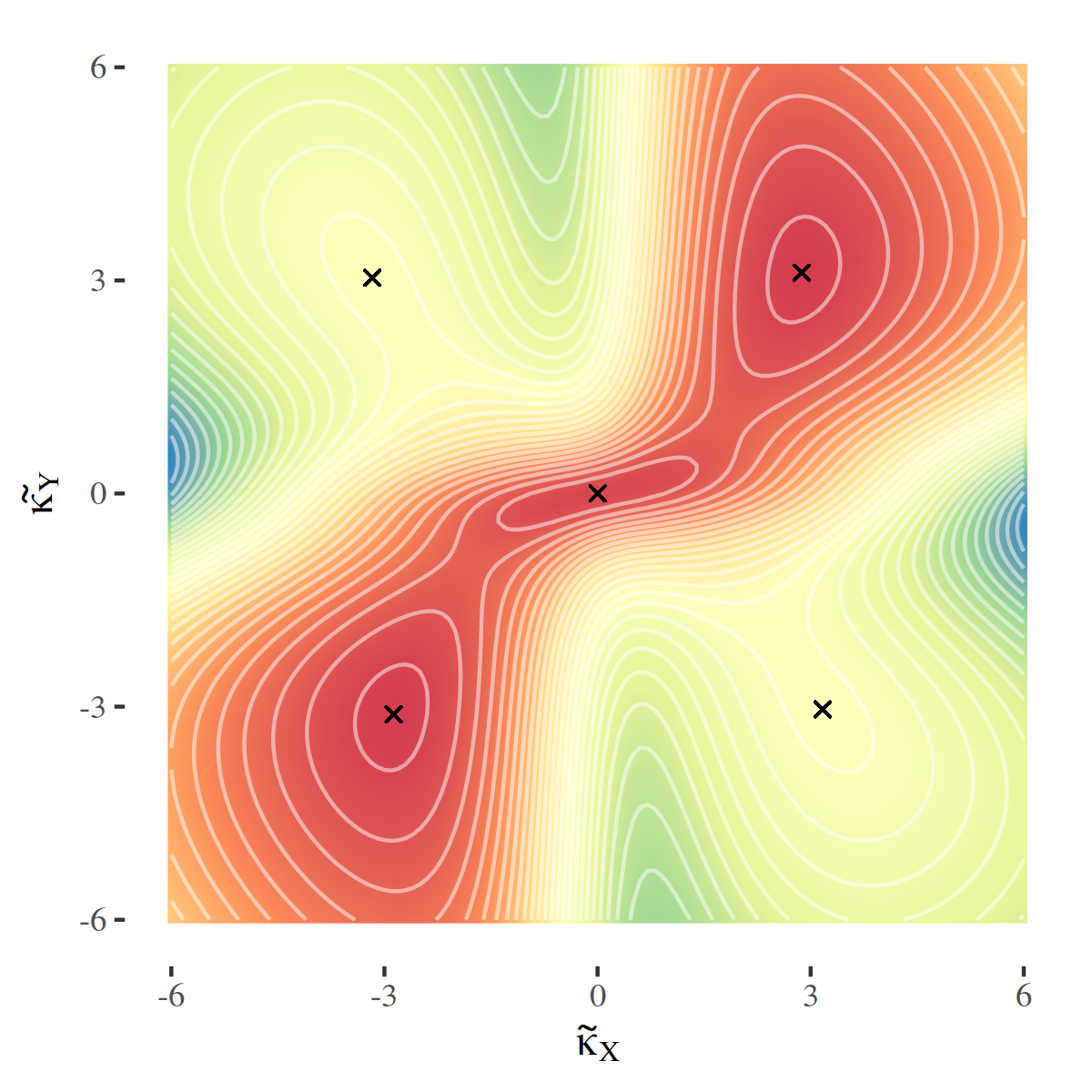}
		\caption{Surface of parameter posterior projected in confounder space, in which five local optima (the black X marks) can be observed.} \label{fig:posterior_contour}
	\end{figure}

	We conjecture that there are at most five posterior local optima for any choice of latent indicator variables, which means that the mixture we intend to use as a posterior approximation will consist of at most five Laplace approximations. We can simplify the optimization by using only three preset initialization points (please see details in supplement) and symmetry. This is typically sufficient to find all the local optima in the full parameter space, or at least the global posterior mode. In Figure~\ref{fig:posterior_contour}, we show an example of posterior surface for which all five local optima are present. The posterior is projected in the confounder space by computing the optimal posterior value for each pair of values $(\scc_\exs, \scc_\out)$. We use the results from the posterior optimization described above to construct an approximation to the posterior density using Laplace's method. We apply the method to each of the (at most five) local optima and then approximate the model evidence by computing the normalization constant for the approximate (unnormalized) posterior, which is a mixture of Laplace approximations (output of Algorithm~\ref{alg:la}). 
	
	\subsection{SAMPLING OVER IV MODELS}
	
	We use the approximated model evidence in a \alg{MC3} scheme to search over the different models. To improve the sampling over causal models, we first run a greedy search procedure to arrive at a good (high-evidence) starting model. The approximations computed during this phase are cached and passed on to the \alg{MC3} stochastic search, after which we prune the explored model list in line with \emph{Occam's window}~\citep{madigan_model_1994} and average over the remaining IV models. By pruning out very-low probabilities estimated models, we arrive at a slimmer, less noisy, and more robust BMA posterior. Finally, we sample the causal effect estimates from the derived BMA posterior distribution. The full set of steps are shown in Algorithm~\ref{alg:massive}. 
	
	\begin{algorithm}[!htb]
		\caption{\alg{MASSIVE} (Model Assessment and Stochastic Search for Instrumental Variable Estimation)}
		\label{alg:massive}
		\begin{algorithmic}
			\STATE {\bfseries Input:} data $\bfZ = [\ivv_i, \exs_i, \out_i]_{1 \le i \le \nobs}$
			\STATE $greedy\_start$ = \alg{greedy\_search($\bfZ$)}
			\STATE $model\_list$ = \alg{MC3\_search}($\bfZ, greedy\_start$)
			\STATE $pruned\_list$ = \alg{prune}($model\_list$)
			\STATE $BMA\_posterior$ = \alg{average($pruned\_list$)}
			\STATE $posterior\_samples$ = \alg{sample($BMA\_posterior$)}
			\STATE {\bfseries Output:} $BMA\_posterior, posterior\_samples$
		\end{algorithmic}
	\end{algorithm}

	\section{EMPIRICAL RESULTS} \label{sec:empirical}
	
	In this experiment we show that our algorithm is accurate in predicting the (lack of) causal effect from $\exs$ to $\out$ when there are least some measured variables that can act as potential instruments. The first and second order statistics for the observed variables $(\ivv, \exs, \out)$ are sufficient statistics for computing the likelihood specified in Equation~\eqref{eqn:likelihood}. If individual-level data is not available, the sufficient statistics can also be derived from summary (regression) data, as shown in the supplement. This means that our approach can leverage the public results obtained from large-sample GWAS. 
	
\begin{figure}[!htb]
		\includegraphics[width=\linewidth]{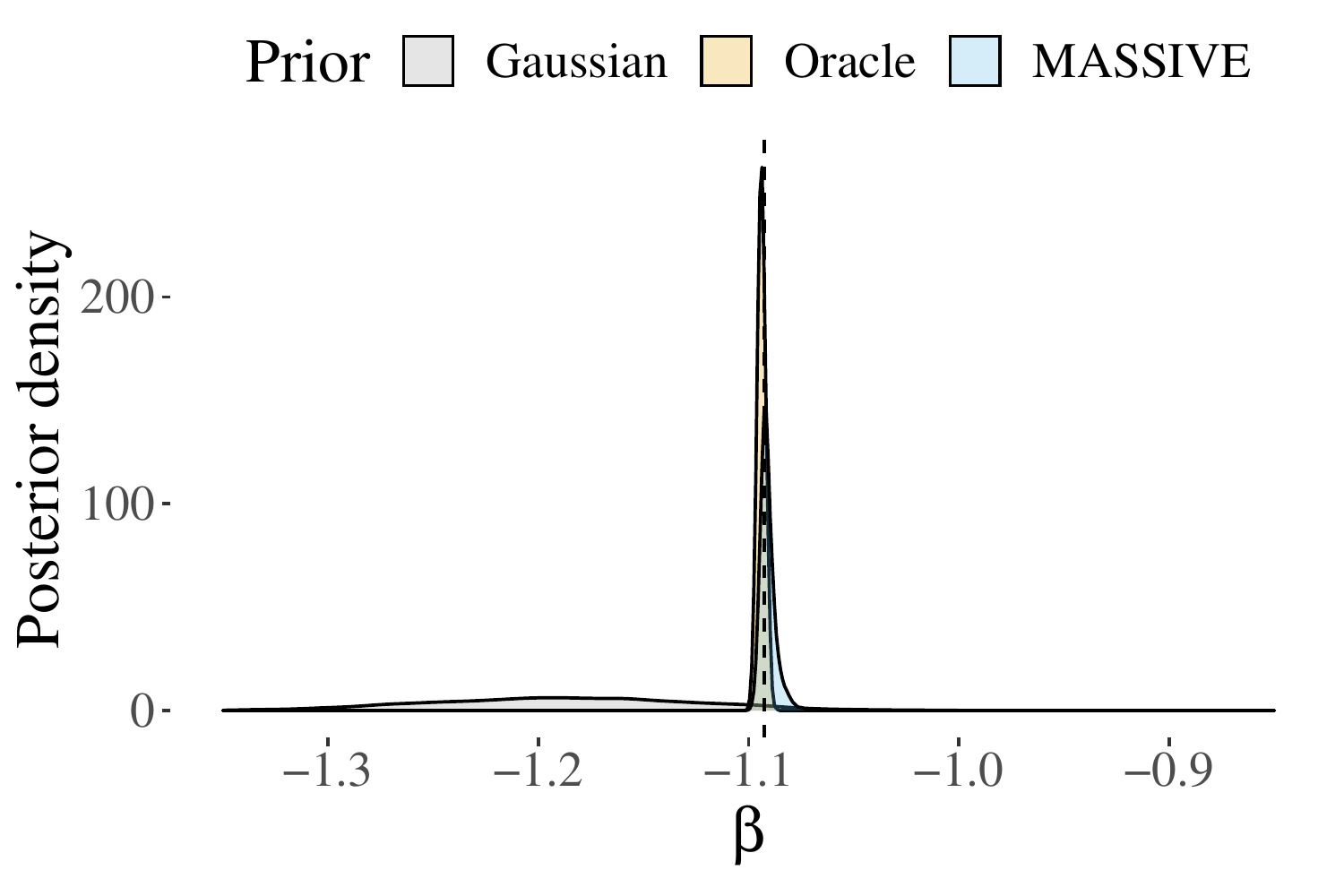}
		\caption{Comparison of estimated causal effect with different sparsifying priors when five out of 50 candidates are valid instruments. The true causal effect value ($\ce = -1.093$) is indicated with a dashed vertical line. \textbf{Gaussian}: We estimate a single model with fixed Gaussian priors on the genetic associations. \textbf{Oracle:} We estimate a single model with a spike-and-slab prior, where the latent indicators on the pleiotropic effects are chosen to correspond to the ground truth, i.e., $\li_j = 0$ if the effect is irrelevant and $\li_j = 1$ if it is relevant. \textbf{MASSIVE:} We use a spike-and-slab prior over the pleiotropic effects and learn the latent indicators with BMA. } \label{fig:compare_priors}
	\end{figure}

	The \emph{selective shrinkage} property of the Gaussian scale mixture leads to an improved causal effect estimate in the scenario under investigation. Without any priors on the pleiotropic effects, the problem is undetermined and for all values of ($\scc_\exs$, $\scc_\out$) we can find a set of parameters that maximizes the data likelihood (please see supplement). By introducing sparsifying priors on the parameters, however, the symmetry among these different sets is broken, leading to a preference for smaller values. The key advantage of the `spike-and-slab' prior is the ability to distinguish between relevant and irrelevant effects. We illustrate this difference in Figure~\ref{fig:compare_priors}. With the spike-and-slab prior, we obtain a much more confident estimate compared to when using a Gaussian prior. In practice, we do not know which of the pleiotropic effects are relevant and which are irrelevant, but with our \alg{MASSIVE} BMA approach, we can infer this distinction from data, thereby significantly improving the causal effect estimate. 
	
	\begin{figure}[!htb]
		\includegraphics[width=\linewidth]{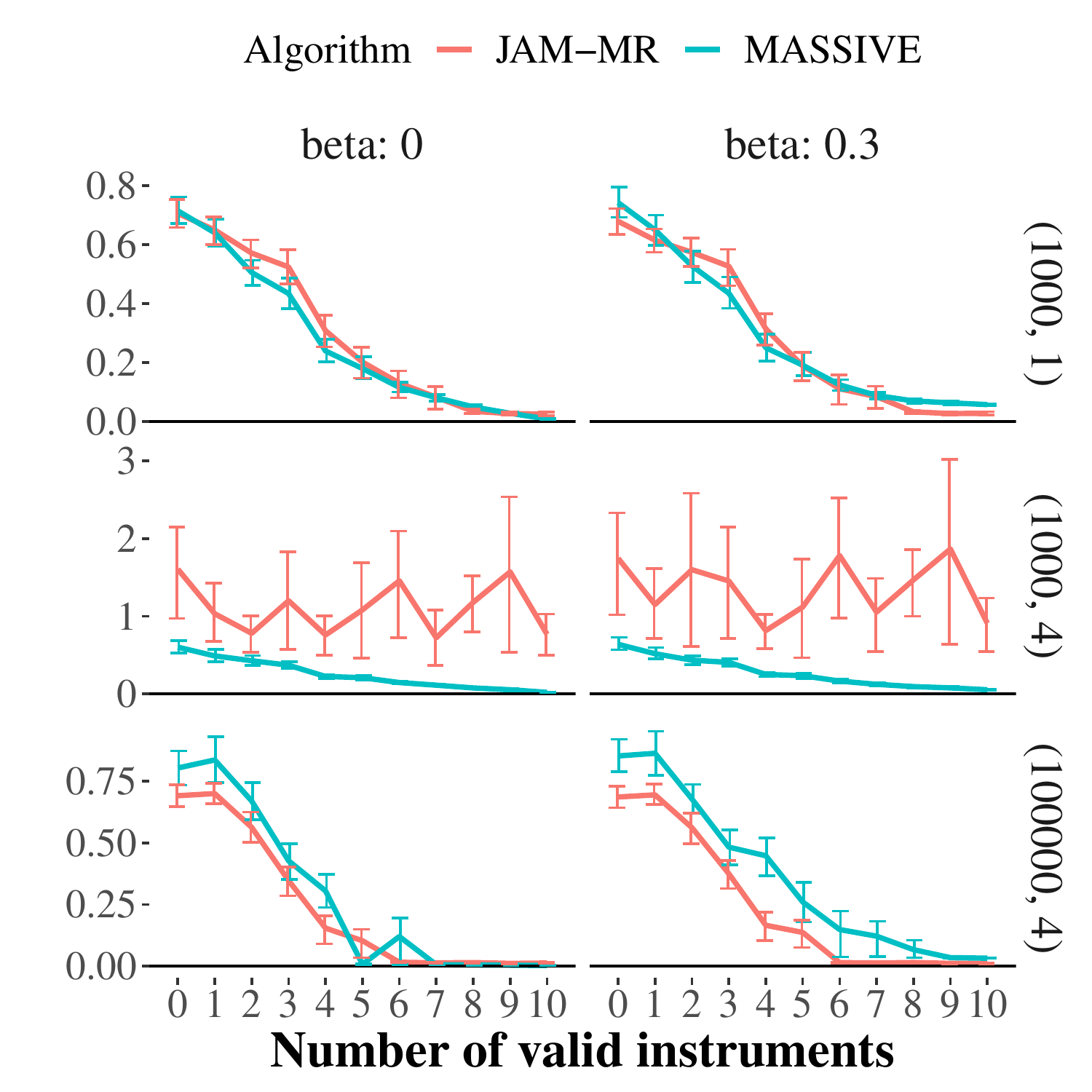}
		\caption{Comparison of \alg{MASSIVE} and \alg{JAM-MR} results averaged over one hundred simulated data sets. \alg{MASSIVE} returns a posterior distribution, unlike \alg{JAM-MR} which outputs point estimates. For \alg{MASSIVE}, we took the median value as the causal point estimate for each data set. We then computed the root mean squared error (RMSE) of the different point causal estimates for both algorithms, as well as the bootstrapped RMSE confidence interval. We ran \alg{JAM-MR} using the default settings, according to which a grid search is used to set the tuning parameter $w$ ~\citep{gkatzionis_bayesian_2019}.} \label{fig:rmse_comparison} 
	\end{figure}
	
	We simulated two different scenarios starting from the setup described in~\citep{gkatzionis_bayesian_2019}: one in which there is no causal effect ($\beta = 0$), and one in which there is a strong positive causal effect ($\beta = 0.3$). The other simulation parameters we varied are the number of generated observations $\nobs$ and the noise $\sigma$, which characterizes the degree of both intrinsic noise and confounding. We considered three simulation configurations: (1) $N = 10^3, \sd = 1$ (less data, less noise); (2) $N = 10^3, \sd = 4$ (less data, more noise); and (3) $N = 10^5, \sd = 4$ (more data, more noise). The full parameters specifications for the linear SEM from Equation~\eqref{eqn:sem} used in the simulated experiments are outlined in~\eqref{eqn:sim_parameters}.  
	\begin{equation}
		\label{eqn:sim_parameters}
		\begin{aligned}
			\nobs \in \{10^3, 10^5\}; J &= 10; \quad K \in \{1, 2, ..., \niv\}; \\
			\forall j \; p_j &\sim \unifd(0.1, 0.9); \\
			\forall j \; \is_j&\sim 0.5 + |\normd(0.0, 0.5^2)|;  \\
			\forall j \; \ply_j &\sim  \pm \mathbf{1}_{j \le K} \normd(0, 0.2^2); \\
			\ce &\in \{0, 0.3\}; \\
			\cc_\exs &= \cc_\out = \sd_\exs = \sd_\out = \sigma \in \{1, 4\}.\\
		\end{aligned}
	\end{equation}
	
	We illustrate the simulation results in Figure~\ref{fig:rmse_comparison}, where we compare our approach against the competing \alg{JAM-MR} algorithm~\citep{gkatzionis_bayesian_2019}. We report the \emph{root mean square error} (RMSE) as a measure of estimation precision. As expected, the estimate generally improves with the number of valid instruments and with noise reduction for both algorithms. In the first configuration, the (potential) instruments are strong, accounting for about 60\% of the variability in $\exs$, while in the other two configurations, they are weak, accounting for around 10\% of the variability. The last configuration is typical for MR studies, which are characterized by large sample sizes but small genetic associations~\citep{davey_smith_mendelian_2014}. Our approach is competitive in the first (less data, less noise) and third (more data, more noise) configuration, and much more robust than $\alg{JAM-MR}$ for the second configuration (less data, more noise).

	\section{REAL-WORLD APPLICATIONS} \label{sec:application}
	
	\subsection{DETERMINANTS OF MACROECONOMIC GROWTH}

	\begin{figure}[!htb]
		\includegraphics[width=\linewidth]{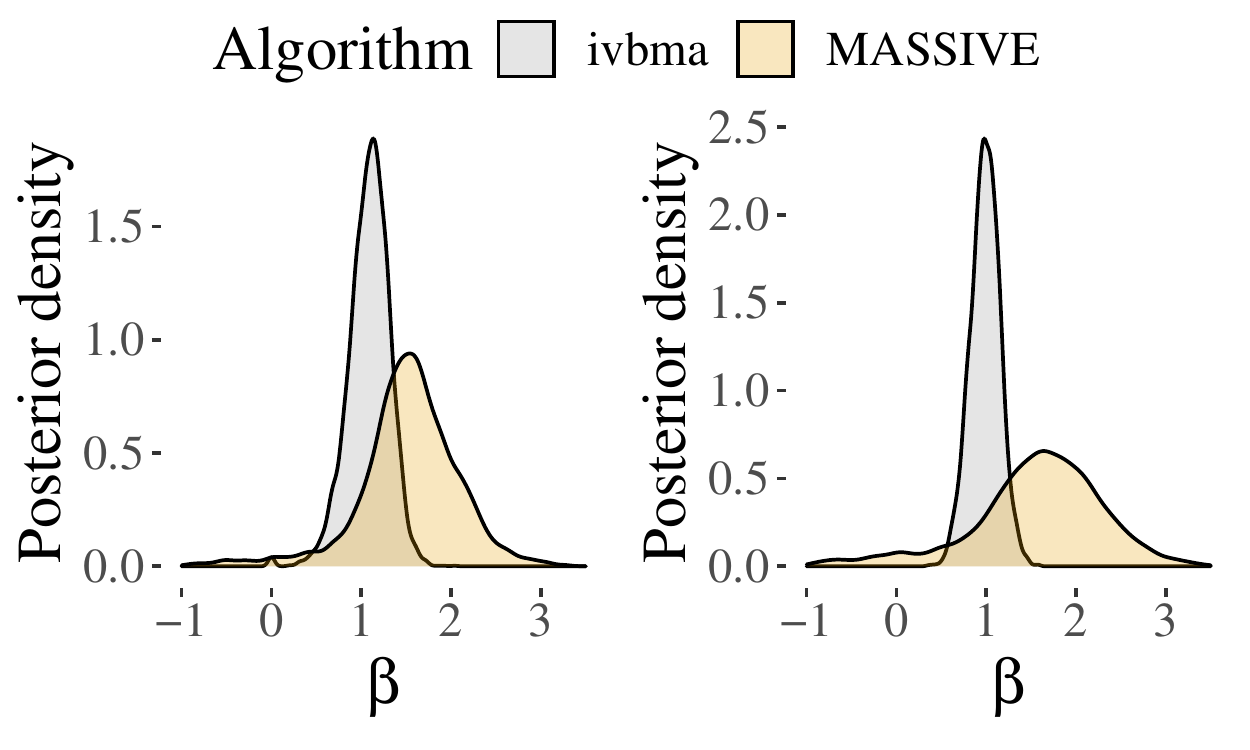}
		\caption{\textbf{Left:} Estimated effect of institutions (rule of law) on macroeconomic growth \textbf{Right:} Estimated effect of economic integration on macroeconomic growth. We used empirically determined values for the \alg{MASSIVE} hyperparameters $\sd_\slab$ and $\sd_\spike$.} \label{fig:growth}
	\end{figure}
	
	In this experiment, we use \alg{MASSIVE} to model uncertainty in macroeconomic growth determinants on a data set compiled by~\citet{rodrik_institutions_2004}. This data set has been previously analyzed by~\citet{karl_instrumental_2012} using the \alg{IVBMA} approach. The goal of the analysis was to find the best determinants (markers) of macroeconomic growth.~\citet{karl_instrumental_2012} found strong evidence indicating \textit{institutions}, as measured by the strength of rule of law, and \textit{economic integration} as the leading determinants of macroeconomic growth. In their analysis, they split the data into the two endogenous variables (exposures), rule of law and integration, four potential instrumental variables and 18 additional covariates. The authors treat these two types of variables distinctly in their model: the instrumental variables are only associated with the exposure, while the covariates are associated with both exposure and outcome. In our model, these two types of variables are considered the same as we do not make any assumptions regarding the candidates' validity a priori, but instead attempt to learn it from the data. Since the \alg{IVBMA} model does not include location parameters, an intercept term is included in the data set, which we also use when running \alg{MASSIVE}. In Figure~\ref{fig:growth} we compare the results obtained with \alg{MASSIVE} and \alg{IVBMA} on the macroeconomic growth data set. The output of \alg{MASSIVE} is in line with previously computed estimates and provides further evidence for a significant causal effect of institutions (rule of law) and economic integration on macroeconomic growth.

	\subsection{INVESTIGATING THE RELATIONSHIP BETWEEN BMI AND PSORIASIS}
	
	\begin{figure}[!htb]
		\includegraphics[width=\linewidth]{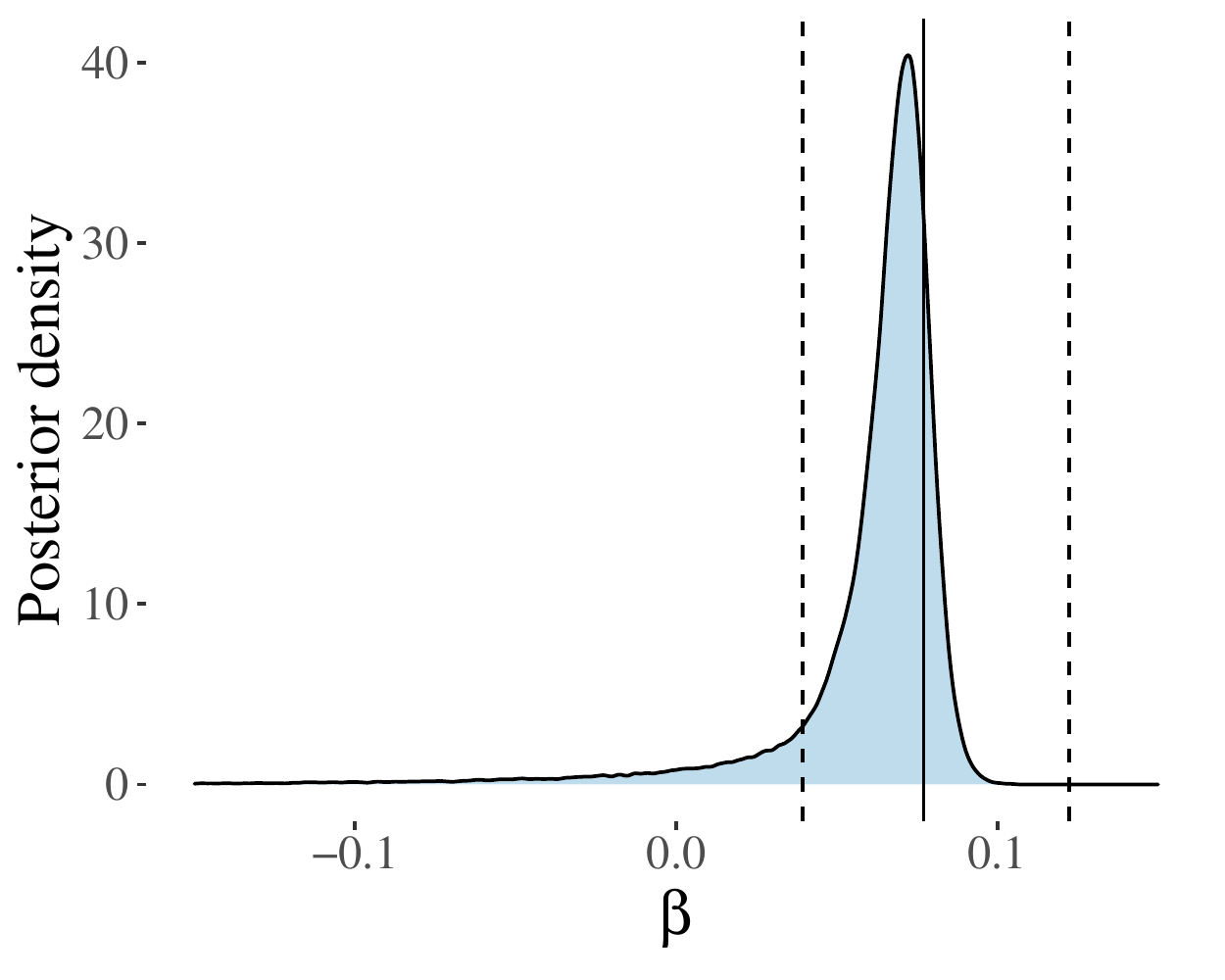}
		\caption{The posterior estimate of the causal effect of BMI, measured in \si{\kilogram\per\meter\squared}, on the log-odds of psoriasis risk obtained when running the \alg{MASSIVE} algorithm with empirically determined hyperparameters. The vertical lines correspond to the results (estimate and error bars) from the IV analysis performed by~\citet{budu-aggrey_evidence_2019} on the UK Biobank data set.} \label{fig:bmi_psoriasis}
	\end{figure}
	
	Psoriasis is a common, chronic inflammatory skin disease, which affects approximately 2-4\% of the population. Psoriasis is presumed to be influenced by both genetic and environmental risk factors, as there are a number of recognized determinants such as family history, smoking, stress, obesity, and alcohol consumption~\citep{parisi_global_2013}.  Establishing a causal link between obesity and psoriasis would be of great clinical interest both for understanding the precise mechanism underlying the association and for guiding treatment recommendations. Recently,~\citet{budu-aggrey_evidence_2019} have attempted to quantify this putative causal relationship by performing an instrumental variable analysis using 97 \textit{single-nucleotide polymorphisms} (SNPs) associated with the \textit{Body Mass Index} (BMI), a common measure of obesity, as genetic instruments. Their study provides evidence that higher BMI leads to a higher risk of psoriasis. The authors report that ``higher BMI causally increased the odds of psoriasis by 9\% per 1 unit increase in BMI". 
	
	In this experiment, we have reproduced their analysis using the \alg{MASSIVE} algorithm. We have applied our approach on the UK Biobank data set analyzed in~\citep{budu-aggrey_evidence_2019}, containing 5,676 psoriasis cases and 372,598 controls. Our algorithm returned 58 models, which were used to compute the model mixture posterior approximation. We then sampled $10^5$ parameter posterior samples from the mixture. In Figure~\ref{fig:bmi_psoriasis} we show the posterior density estimate for the causal effect $\ce$. The result obtained is very similar to that reported in~\citep{budu-aggrey_evidence_2019}. It provides further evidence for increased BMI leading to a higher occurrence of psoriasis.

	\section{DISCUSSION}
	
	It is crucial to take model uncertainty into account when making inferences so as to mitigate the pitfalls of model misspecification~\citep{hoeting_bayesian_1999}. Bayesian Model Averaging (BMA) is a principled approach of incorporating this uncertainty into the analysis, but it is limited in scope due to the intractability of evaluating the model evidence for a considerable number of interesting models. In light of the computational limitations, the researcher often turns to approximating the evidence, but common solutions such as the BIC approximation might not be suitable for complex models~\citep{fragoso_bayesian_2018}. Through a combination of clever model choices and a hybrid inference scheme, combining \alg{MC3} stochastic search with fast Laplace approximations, \alg{MASSIVE} is the first algorithm that can provide a reliable posterior estimate of the causal effect in IV settings with hundreds of candidate instruments. 
	
	Our proposed model provides a flexible and general solution for instrumental variable analyses. Thanks to the `spike-and-slab' type prior on the interaction strengths, potential background knowledge regarding the sparsity and effective size of interactions can easily be incorporated into the model in an intuitive fashion. In this work, we have chosen to model the confounding coefficients explicitly in order to provide a unified view of causal interactions. Another possibility would have been to model the confounding effect as variance terms in a correlated errors model~\citep{jones_choice_2012}, a possibility we leave for future work. 
	
	In our Bayesian approach, we have proposed simple but flexible priors both over the model and parameter space to permit a more accurate approximation of the posterior using Laplace's method. This approach allows for a tractable search through the model space, and parameter samples can be immediately derived from the approximation. The approach also lends itself to straightforward parallelization. In future work we plan to refine and speed up the process by, for example, including more starting points in the optimization procedure and distributing them across multiple cores. Furthermore, there is great potential in combining our approach with other means of (pre-)selecting instruments such as Lasso-based methods~\citep{belloni_sparse_2012} or the sparse IV (\alg{SPIV}) approach ~\citep{agakov_sparse_2010}.

	\subsubsection*{Acknowledgements}
	
	This research has been partially financed by the Netherlands Organisation for Scientific Research (NWO) under project 617.001.451 (CHiLL). 
	
	
	\printbibliography[heading=uai]
	
	\appendix
	
	\section{DERIVING THE MAXIMUM LIKELIHOOD ESTIMATES} \label{sec:derive_mle}
	
	In our model, we assume that the data is generated from the following linear structural equation model:
	\begin{equation}
		\label{eqn:sem}
		\begin{aligned}
			\conf &:= \err_\conf \\
			\iv_j &:= \err_{\iv_j} \\
			\exs &:= \sum_j \is_j \iv_j + \cc_\exs \conf + \err_\exs \\ 
			\out &:= \sum_j \ply_j \iv_j + \cc_\out \conf + \ce \exs + \err_\out
		\end{aligned}.
	\end{equation}
	
	In order to avoid any scaling issues, we first divide each structural equation in~\eqref{eqn:sem} by the scale of the noise term. We then define priors on the scale-free interactions. The scaled structural parameters are
	\begin{align*}
		\sis_j &= \sd_{\iv_j} \sd_\exs^{-1} \is_j; \quad \sply_j = \sd_{\iv_j} \sd_\out^{-1} \ply_j; \\
		\sce &= \sd_\exs \sd_\out^{-1} \ce; \quad \scc_\exs = \sd_\exs^{-1} \cc_\exs; \quad \scc_\out = \sd_\out^{-1} \cc_\out. \\
	\end{align*}
	
	We assume that the data consists of $\nobs$ i.i.d. observations $\data = (\ivv_i, \exs_i, \out_i)_{1 \le i \le \nobs}$. The conditional Gaussian observed data likelihood reads
	\begin{equation} \label{eqn:likelihood_appendix}
		\lik \left( \left. \begin{bmatrix} \exs \\ \out \end{bmatrix} \right| \ivv \right) = (4\pi^2|\bSigma|)^{-\frac{\nobs}{2}} \exp\left\{-  \frac{\nobs}{2} \textrm{tr}(\bSigma^{-1} \bfS) \right\}, 
	\end{equation}
	with $$\bfS = \by{\nobs} \sum\limits_{i = 1}^\nobs \left\{\begin{bmatrix} \exs_i \\ \out_i \end{bmatrix} - \bmu(\ivv_i) \right\} \left\{\begin{bmatrix} \exs_i \\ \out_i \end{bmatrix} - \bmu(\ivv_i) \right\}^\trp$$ and 
	$$\bmu(\ivv) = \begin{bmatrix} \isv & \ce \isv + \plyv \end{bmatrix}^\trp \ivv = \begin{bmatrix} \isv & \tev \end{bmatrix}^\trp \ivv.$$
	
	The maximum of the conditional likelihood function occurs at $\bfS = \bSigma$. Our model has $2\niv + 5$ (scaled) parameters, $\sparm = (\sisv, \splyv, \sce, \log\sd_\exs, \log\sd_\out, \scc_\exs, \scc_\out)$, which is more than the number of independent constraints ($2\niv + 3$) imposed by maximizing the likelihood. This makes the problem of finding the maximum likelihood estimate undetermined, but if we fix the values of $\scnfm = (\scc_\exs, \scc_\out)$, we can analytically derive the other parameters ($\sobsm$) such that the likelihood is maximized.
	
	We have as input sufficient statistics the first and second-order empirical (raw) moments of the data:
	\begin{align*}
		\overline{\ivv} &= \by{N} \sum_{i=1}^N \ivv_i \to \EV{\ivv}; \\
		\overline{\exs} &= \by{N} \sum_{i=1}^N \exs_i \to \EV{\exs}; \\
		\overline{\out} &= \by{N} \sum_{i=1}^N \out_i \to \EV{\out}; \\
		\overline{\ivv \ivv^\trp} &= \by{N} \sum_{i=1}^N \ivv_i \ivv^\trp_i \to \EV{\ivv \ivv^\trp}; \\
		\overline{\ivv \exs} &= \by{N} \sum_{i=1}^N \ivv_i \exs_i \to \EV{\ivv \exs}; \\
		\overline{\ivv \out} &= \by{N} \sum_{i=1}^N \ivv_i \out_i \to \EV{\ivv \out}; \\
		\overline{\exs^2} &= \by{N} \sum_{i=1}^N \exs_i^2 \to \EV{\exs^2}; \\
		\overline{\out^2} &= \by{N} \sum_{i=1}^N \out_i^2 \to \EV{\out^2}; \\
		\overline{\exs \out} &= \by{N} \sum_{i=1}^N \exs_i \out_i \to \EV{\exs \out}. \\
	\end{align*}
	
	The maximum likelihood estimator here coincides with the method of moments estimator, so we will derive the ML estimates using moment matching, which is straightforward. The conditional moments relate to the parameters as follows:
	\begin{align*}
		\EV{\exs \given \ivv} &= \isv^\trp \ivv \\
		\EV{\out \given \ivv} &= \tev^\trp \ivv \\
		\Var{\exs \given \ivv} &= \sd_\exs^2 + \cc_\exs^2 \\
		\Cov{\exs, \out \given \ivv} &= \ce (\sd_\exs^2 + \cc_\exs^2) + \cc_\exs \cc_\out \\
		\Var{\out \given \ivv} &= \sd_\out^2 + \ce^2 \sd_\exs^2 + (\cc_\out + \ce \cc_\exs)^2.
	\end{align*}
	
	We now relate the previous statements to the unconditional moments:
	\begin{align*}
		\EV{\ivv \exs} &= \EV{\ivv \EV{\exs \given \ivv}} = \EV{\ivv \ivv^\trp} \isv \\
		\EV{\ivv \out} &= \EV{\ivv \EV{\out \given \ivv}} = \EV{\ivv \ivv^\trp} \tev \\
		\EV{\exs^2} &= \isv^\trp \EV{\ivv \ivv^\trp} \isv + \sd_\exs^2 + \cc_\exs^2 \\
		\EV{\exs \out} &= \isv^\trp \EV{\ivv \ivv^\trp} \tev + \ce (\sd_\exs^2 + \cc_\exs^2) + \cc_\exs \cc_\out \\
		\EV{\out^2} &= \tev^\trp \EV{\ivv \ivv^\trp} \tev + \sd_\out^2 + \ce^2 \sd_\exs^2 + (\cc_\out + \ce \cc_\exs)^2.
	\end{align*}
	We therefore obtain the constraints
	{\small
		\begin{align*}
			\isv &= (\overline{\ivv \ivv^\trp})^{-1} \overline{\ivv \exs} \\
			\ce \isv + \plyv &= (\overline{\ivv \ivv^\trp})^{-1} \overline{\ivv \out} \\
			\sd_\exs^2 + \cc_\exs^2 &= \sVar{\exs \given \ivv}\\ 
			&= \overline{\exs^2} - \overline{\exs \ivv^\trp} (\overline{\ivv \ivv^\trp})^{-1} \overline{\ivv \exs} \\
			\ce (\sd_\exs^2 + \cc_\exs^2) + \cc_\exs \cc_\out &= \sCov{\exs, \out \given \ivv} \\
			&= \overline{\exs\out} - \overline{\exs \ivv^\trp} (\overline{\ivv \ivv^\trp})^{-1} \overline{\ivv \out} \\
			\sd_\out^2 + \ce^2 \sd_\exs^2 + (\cc_\out + \ce \cc_\exs)^2 &= \sVar{\out \given \ivv} \\
			&= \overline{\out^2} - \overline{\out \ivv^\trp} (\overline{\ivv \ivv^\trp})^{-1} \overline{\ivv \out}.
		\end{align*}
	}%
	The next step is to express the above constraints in terms of the scaled parameters:
	{\small
		\begin{align*}
			\Var{\ivv} \sisv \sd_\exs^{-1} &= (\overline{\ivv \ivv^\trp})^{-1} \overline{\ivv \exs} \\
			\Var{\ivv} (\sce \sisv + \splyv) \sd_\out^{-1} &= (\overline{\ivv \ivv^\trp})^{-1} \overline{\ivv \out} \\
			\sd_\exs^2 (1 + \scc_\exs^2) &= \sVar{\exs \given \ivv}\\ 
			&= \overline{\exs^2} - \overline{\exs \ivv^\trp} (\overline{\ivv \ivv^\trp})^{-1} \overline{\ivv \exs} \\
			\sd_\exs \sd_\out \left[ \sce (1 + \scc_\exs^2) + \scc_\exs \scc_\out \right] &= \sCov{\exs, \out \given \ivv} \\
			&= \overline{\exs\out} - \overline{\exs \ivv^\trp} (\overline{\ivv \ivv^\trp})^{-1} \overline{\ivv \out} \\
			\sd_\out^2 \left[ 1 + \sce^2 + (\scc_\out + \sce \scc_\exs)^2 \right] &= \sVar{\out \given \ivv} \\
			&= \overline{\out^2} - \overline{\out \ivv^\trp} (\overline{\ivv \ivv^\trp})^{-1} \overline{\ivv \out}.
		\end{align*}
	}%
	From the above constraints, given fixed values for $\scc_\exs$ and $\scc_\out$, we obtain the following (scaled) parameter values that maximize the likelihood in~\eqref{eqn:likelihood_appendix}:
	
	{\small
		\begin{equation} \label{eqn:mle}
			\begin{aligned}
				(\sd_\exs^\ML)^2 &= \frac{\sVar{\exs \given \ivv}}{1 + \scc_\exs^2} \\
				(\sd_\out^\ML)^2 &= \left(\sVar{\out \given \ivv} - \frac{ (\sCov{\exs, \out \given \ivv})^2}{\sVar{\exs \given \ivv}}\right) \frac{1 + \scc_\exs^2}{1 + \scc_\exs^2 + \scc_\out^2} \\
				\sce^\ML &= \frac{\sCov{\exs, \out \given \ivv} (\sd_\exs^\ML \sd_\out^\ML)^{-1} - \scc_\exs \scc_\out}{1 + \scc_\exs^2}  \\
				\sisv^\ML &= \sqrt{\sVar{\ivv}} (\sEV{\ivv\ivv^\trp})^{-1} \sEV{\ivv \exs} (\sd_\exs^\ML)^{-1} \\
				\splyv^\ML &= \sqrt{\sVar{\ivv}} (\sEV{\ivv\ivv^\trp})^{-1} \sEV{\ivv \out} (\sd_\out^\ML)^{-1} - \sce^\ML \sisv^\ML
			\end{aligned}.
		\end{equation}
	}
	
	\section{USING SUMMARY STATISTICS AS INPUT}
	
	To run \alg{MASSIVE}, we must provide the first and second-order moments of the observed data $\bfZ = (\ivv, \exs, \out)$ as input to plug into the data likelihood from Equation~\eqref{eqn:likelihood_appendix}. If we have access to the whole data set, then the moments can immediately be derived. Much more often, however, individual-level data is unavailable and instead we have to rely on published GWAS results, which typically come in the form of regression coefficients together with their standard errors. In this section we show how the first and second-order moments can be derived from this summary data, thereby making \alg{MASSIVE} applicable on a much broader set of data sources.
	
	To obtain all the necessary input sufficient statistics, we require the following summary data:
	\begin{itemize}
		\item $\hat{\eaf}_j$: the effect allele frequency (EAF) of $\iv_j$
		\item $\nac$: the number of allele copies (almost always equal to two, since humans are diploid organisms)
		\item $\hat{\is}_j, \hat{\sigma}_{\hat{\is}_j}, \nobs_{\hat{\is}_j}$: for the gene-exposure associations, we require the coefficient obtained by regressing $\exs$ on $\iv_j$ , its standard error and the sample size 
		\item $\hat{\te}_j, \hat{\sigma}_{\hat{\te}_j}, \nobs_{\hat{\te}_j}$: for the gene-outcome associations, we require the coefficient obtained by regressing $\out$ on $\iv_j$, its standard error and the sample size
		\item $\hat{\ce}$: the coefficient obtained by regressing $\exs$ on $\out$ (observational exposure-outcome association)
	\end{itemize}
	
	Summary data on gene-exposure and gene-outcome associations from GWAS is widely available, so we can typically get estimates for $\hat{\is}_j$, $\hat{\te}_j$ together with the associated standard errors and sample sizes. The effect allele frequency $\hat{\eaf}_j$ is usually also reported. In addition, we require a measure of the association between the exposure and the outcome ($\hat{\ce}$) to derive an estimate of $\Cov{\exs, \out}$. This estimate can be obtained from observational studies for determining potential risk factors for the outcome. 
	
	To estimate the second-order moments, we employ the following well-known approximations from simple linear regression:
	\begin{equation*}
		\small
		\begin{aligned}
			\hat{\is}_j &\approx \frac{\Cov{\iv_j, \exs}}{\Var{\iv_j}} \\
			\hat{\te}_j &\approx \frac{\Cov{\iv_j, \out}}{\Var{\iv_j}} \\
			\hat{\ce} &\approx \frac{\Cov{\exs, \out}}{\Var{\exs}} \\
			\hat{\sigma}^2_{\hat{\is}_j} &\approx \frac{1}{N_\is} \left( \frac{\Var{\exs}}{\Var{\iv_j}} - \hat{\is}^2 \right) \\
			\hat{\sigma}^2_{\hat{\te}_j} &\approx \frac{1}{N_\te} \left( \frac{\Var{\out}}{\Var{\iv_j}} - \hat{\te}^2 \right)
		\end{aligned}.
	\end{equation*}
	
	Note that these approximations also apply in a multivariate setting when the regressors are independent. Moreover, to compute the expected values and variances for the genetic variants, we assume a binomial distribution, so we plug in the EAF as the estimated success probability and then use the appropriate formulas.  We use all these approximations to finally derive the following estimates for the moments from summary statistics:
	
	\begin{equation} 
		\small \label{eqn:summary_stat}
		\begin{aligned} 
			\EV{\iv_j} &\approx \nac \cdot \hat{\eaf}_j \; (= \widehat{\EV{\iv_j}}) \\
			\EV{X} &\approx \sum_j \widehat{\EV{\iv_j}} \cdot \hat{\is}_j \\
			\EV{Y} &\approx \sum_j \widehat{\EV{\iv_j}} \cdot \hat{\te}_j \\
			\Var{\iv_j} &\approx \nac \cdot \hat{\eaf}_j \cdot (1 - \hat{\eaf}_j) \; (= \widehat{\Var{\iv_j}}) \\
			\Cov{\iv_j, \exs} &\approx \widehat{\Var{\iv_j}} \cdot \hat{\is}_j \\
			\Cov{\iv_j, \out} &\approx \widehat{\Var{\iv_j}} \cdot \hat{\te}_j \\
			\Var{\exs} &\approx \widehat{\Var{\iv_j}} \cdot (\hat{\is}_j^2 + \nobs_{\hat{\is}_j} \cdot \hat{\sigma}^2_{\hat{\is}_j}) \; (= \widehat{\Var{\exs}}) \\
			\Var{\out} &\approx \widehat{\Var{\iv_j}} \cdot (\hat{\te}_j^2 + \nobs_{\hat{\te}_j} \cdot \hat{\sigma}^2_{\hat{\te}_j}) \\
			\Cov{\exs, \out} &\approx \widehat{\Var{\exs}} \cdot \hat{\ce}
		\end{aligned}.
	\end{equation}
	When we have information on multiple genetic variants, we obtain multiple estimates of $\Var{\exs}$ and $\Var{\out}$ in~\eqref{eqn:summary_stat}, in which case we take the median over the estimates. Our approach also requires specifying a sample size. Since the summary statistics are likely to be computed from different samples, we conservatively choose the minimum of their sizes as input to \alg{MASSIVE} in order not to overestimate the precision of the data. If the sample size for the exposure-outcome association measure is also available, we take it into consideration when calculating the minimum of the sample sizes.
	
	\section{SMART INITIALIZATION PROCEDURE FOR THE POSTERIOR OPTIMIZATION}
	
	We propose to start the search for posterior local optima from the bivariate maximum likelihood manifold. Since we are looking for sparse parameter solutions, we also start from points on the manifold that exhibit some degree of sparsity.
	
	The first starting point corresponds to the \emph{no confounding} sparse solution, where we fix $\scc_\exs = \scc_\out = 0$. The other parameters can be derived using~\eqref{eqn:mle}:
	{\small
		\begin{align*}
			(\sd_\exs^\ML)^2 &= \sVar{\exs \given \ivv} \\
			(\sd_\out^\ML)^2 &= \sVar{\out \given \ivv} - \frac{ (\sCov{\exs, \out \given \ivv})^2}{\sVar{\exs \given \ivv}} \\
			\sce^\ML &= \sCov{\exs, \out \given \ivv} (\sd_\exs^\ML \sd_\out^\ML)^{-1}   \\
			\sisv^\ML &= \sVar{\ivv} (\sEV{\ivv\ivv^\trp})^{-1} \sEV{\ivv \exs} (\sd_\exs^\ML)^{-1} \\
			\splyv^\ML &= \sVar{\ivv} (\sEV{\ivv\ivv^\trp})^{-1} \sEV{\ivv \out} (\sd_\out^\ML)^{-1} - \sce^\ML \sisv^\ML.
		\end{align*}
	}%
	
	The second starting point corresponds to the \emph{no causal effect} solution, where we fix $\sce = 0$. By solving the equation system in~\eqref{eqn:mle} with $\sce = 0$, we obtain the following constraint:
	$ \scc_\exs \scc_\out = \frac{\sCov{\exs, \out \given \ivv}}{1 - \sCov{\exs, \out \given \ivv}}.$
	
	We have one degree of freedom left for choosing $\scc_\exs$ and $\scc_\out$. We propose to additionally set $|\scc_\exs| = |\scc_\out|$ and assume $\scc_\exs > 0$. Finally, we obtain:
	
	{\small
		\begin{align*}
			\scc_\exs^\ML &= \sqrt{\left|\frac{\sCov{\exs, \out \given \ivv}}{1 - \sCov{\exs, \out \given \ivv}}\right|} \\
			\scc_\out^\ML &= \sqrt{\left|\frac{\sCov{\exs, \out \given \ivv}}{1 - \sCov{\exs, \out \given \ivv}}\right|} \cdot \sign \left\{\frac{\sCov{\exs, \out \given \ivv}}{1 - \sCov{\exs, \out \given \ivv}}\right\}.
		\end{align*}
	}%
	The rest of the parameters can be derived given $(\scc_\exs, \scc_\out)$ from~\eqref{eqn:mle}. 
	
	The third starting point corresponds to minimizing the pleiotropic effects sum of squares. If we consider the constraint (at the maximum likelihood estimate):
	$$ \plyv = (\overline{\ivv \ivv^\trp})^{-1} \overline{\ivv \out} - \ce (\overline{\ivv \ivv^\trp})^{-1} \overline{\ivv \exs} = r^{\out \given \ivv} - \ce r^{\exs \given \ivv},$$
	where $r^{\exs \given \ivv}$ and $r^{\out \given \ivv}$ are the coefficients obtained by regressing $\ivv$ on $\exs$ and $\out$, respectively, 
	$$\ce^* = \arg\min \sum_{j=1}^J \ply_j^2 = \arg\min \left( r_j^{\out \given \ivv} - \ce r_j^{\exs \given \ivv} \right)^2.$$
	The solution to this minimization problem is:
	$$ \ce^* = \frac{1}{\niv} \frac{\sum_{j=1}^\niv  r_j^{\exs \given \ivv} r_j^{\out \given \ivv}}{\sum_{j=1}^\niv  r_j^{\exs \given \ivv} r_j^{\exs \given \ivv}}.$$
	For independent instruments, the right-hand side ratios above corresponds to the instrumental variable estimates.  By solving the equation system in~\eqref{eqn:mle} with $\ce = \ce^*$, we obtain the following constraint:
	$ \scc_\exs \scc_\out = \frac{C}{1 - C},$
	where 
	{\scriptsize
		\begin{align*} 
			C &= \sCov{\exs, \out \given \ivv} - \\
			&-\frac{\ce^* \sVar{\exs \given \ivv}}{
				\sqrt{\sVar{\exs \given \ivv} (\sVar{\out \given \ivv} + (\ce^*)^2 \sVar{\exs \given \ivv} - 2 \ce^* \sCov{\exs, \out \given \ivv})}}.
		\end{align*}
	}%
	
	We have one degree of freedom left for choosing $\scc_\exs$ and $\scc_\out$. We propose to additionally set $|\scc_\exs| = |\scc_\out|$ and assume $\scc_\exs > 0$. Finally, we obtain:
	\begin{align*}
		\small
		\scc_\exs^\ML &= \sqrt{\left|\frac{C}{1 - C}\right|} \\
		\scc_\out^\ML &= \sqrt{\left|\frac{C}{1 - C}\right|} \cdot \sign \left\{\frac{C}{1 - C}\right\}. \\
	\end{align*}

	\section{DETERMINING THE PRIOR HYPERPARAMETERS EMPIRICALLY}
	
	We base our choice of prior hyperparameters on how likely it is for the observed genetic associations to have come from the prior. We start by choosing a reasonable hyperparameter for the `slab' component ($\sd_\slab$). Since instrument candidates are chosen based on the robustness of their association with the exposure $\exs$, we can use the size of these associations as a measure of the effect size of relevant effects, i.e., those corresponding to the `slab' component. In our framework, this translates to the assumption that all the instrument strengths $\sis_j$ arise from the `slab' distribution, and will therefore give a good indication of the expected effect size for relevant parameters. Consequently, we want to find the hyperparameter value that maximizes the (log-)likelihood of the genetic associations coming from $\normd(0, \sd_\slab^2)$:
	\begin{equation} \label{eqn:min_slab}
		\sd_\slab^* = \argmax_{\sd_\slab} \sum_{j=1}^\niv \left[ - \frac{1}{2} \log(2\pi \sd_\slab^2) - \frac{\sis_j^2}{2 \sd_\slab^2} \right].
	\end{equation}
	
	Maximizing the above log-likelihood is straightforward if we know the instrument strengths $\sis_j$ on the right-hand side from data. Instead, we will plug in an empirical estimate of the scaled instrument strengths. We use the fact that the unscaled maximum likelihood estimate for the instrument strengths $\is_j$ is identifiable as
	$$ \isv^\ML = (\overline{\ivv\ivv^\trp})^{-1} \overline{\ivv\exs}. $$
	
	For the scaled parameters we then have:
	$$ (\sis_j^\ML)^2 = \sd_{\iv_j}^2 (\is_j^\ML)^2 (\sd_\exs^\ML)^{-2} = \frac{\sd_{\iv_j}^2 (\is_j^\ML)^2 (1 + \scc_\exs^2)}{\Var{\exs \given \ivv}}.$$
	
	These values are undetermined because we de not know the confounding coefficient $\scc_\exs$. We propose to compute an average estimate by integrating out $\scc_\exs$, which we have assumed follows a $\normd(0, 10)$ distribution a-priori. We average over all possible values of $\scc_\exs$ to get
	{\small
		\begin{equation} \label{eqn:sis_estimate}
			\begin{aligned}
				\EV{(\sis_j^\ML)^2} &= \frac{\sd_{\iv_j}^2 (\is_j^\ML)^2}{\Var{\exs \given \ivv}} \int_{-\infty}^{\infty} (1 + \scc_\exs^2) \; \normd(\scc_\exs; 0, 10) \diff\scc_\exs \\
				&= \frac{\sd_{\iv_j}^2 (\is_j^\ML)^2}{\Var{\exs \given \ivv}} \cdot 101 \\
				&\overset{!}{=} 101 D_j^2.
			\end{aligned}
		\end{equation}
	}%
	
	We plug in the derived estimate into~\eqref{eqn:min_slab} to get
	$$ 	\sd_\slab^* = \argmax_{\sd_\slab} \sum_{j=1}^\niv \left[ - \log \sd_\slab - \frac{101 D_j^2}{2 \sd_\slab^2} \right].$$
	
	From this we finally obtain our first empirically determined hyperparameter
	\begin{equation} \label{eqn:sd_slab_dh}
		(\sd_\slab^*)^2 = \frac{101}{\niv} \sum_{j=1}^\niv D_j^2 =   \frac{101}{\niv} \sum_{j=1}^\niv \frac{\sd_{\iv_j}^2 (\is_j^\ML)^2}{\Var{\exs \given \ivv}}.
	\end{equation}
	
	We now derive a reasonable hyperparameter for the `spike' component ($\sd_\spike$), relative to the previously determined $\sd_\slab^*$. The potential gain (or penalty) in moving $\sis_\textrm{min}$ from the slab to the spike component in the prior is
	$$ G(\sd_\spike, \sd_\slab) = \log\normd(\sis_\textrm{min}; 0, \sd_\spike) - \log\normd(\sis_\textrm{min}; 0, \sd_\slab).$$
	The penalty in the likelihood (approximated by a normal distribution) due to the parameter shrinkage from its current value to zero is
	$$ P(\sd_\slab) = \nobs \cdot [ \log\normd(0; \sis_\textrm{min}, \sd_\slab) - \log\normd(\sis_\textrm{min}; \sis_\textrm{min}, \sd_\slab) ].$$
	
	The empirical argument we employ is to choose $\sd_\spike$ so small such that changing the component of the minimal instrument strength ($\sis_\textrm{min} = \min_j \sis_j$) from slab to spike would incur a greater penalty than the one induced on the log-likelihood by shrinking that parameter to zero. This way, fitting any $\sis_j$ into the `spike' component is strongly discouraged, in line with our assumption that these are relevant values coming from the `slab' component. Consequently, as our second empirically determined hyperparameter, we choose the value $\sd_\spike^*$ solving the equation $G(\sd_\spike, \sd_\slab^*) = P(\sd_\slab^*)$, where $\sd_\slab^*$ is given in~\eqref{eqn:sd_slab_dh} and our estimate of $\sis_\textrm{min}$ is the smallest of the $\niv$ expected value estimates derived in~\eqref{eqn:sis_estimate}. It is straightforward to show that the constraint boils down to
	$$ (\nobs + 1 - C) \left(\frac{101 \min_j D_j^2}{\sd_\slab^*}\right)^2 + \log{C} = 0,$$
	where $C = \left(\frac{\sd_\slab^*}{\sd_\spike}\right)^2$. It can be easily shown that the above equation in $C$ has a unique solution greater than one ($C > 1$ by definition because $\sd_\spike < \sd_\slab$).  Via our empirical argument, we have thus arrived at an easily computable, unique pair of hyperparameters ($\sd_\slab^*, \sd_\spike^*$). We emphasize that this choice of parameters is independent of the true causal effect and relies solely on the estimated values of the instrument strengths to calibrate the appropriate size of relevant (`slab') and irrelevant (`spike') effects.

\end{document}